\documentclass[conference, a4paper]{IEEEtran}
\IEEEoverridecommandlockouts
\usepackage{cite}
\usepackage{amsmath,amssymb,amsfonts}
\usepackage{algorithmic}
\usepackage{graphicx}
\usepackage{textcomp}
\usepackage{xcolor}
\def\BibTeX{{\rm B\kern-.05em{\sc i\kern-.025em b}\kern-.08em
    T\kern-.1667em\lower.7ex\hbox{E}\kern-.125emX}}
\usepackage{booktabs}
\usepackage{graphicx}
\usepackage{amsmath}
\usepackage{amssymb}
\usepackage{booktabs}
\usepackage{blindtext}
\usepackage{hyperref}
\usepackage{comment}
\usepackage{subcaption}
\usepackage{xspace}
\usepackage{xurl}

\usepackage[capitalize]{cleveref}
\crefname{section}{Sec.}{Secs.}
\Crefname{section}{Section}{Sections}
\Crefname{table}{Table}{Tables}
\crefname{table}{Tab.}{Tabs.}

\makeatletter
\DeclareRobustCommand\onedot{\futurelet\@let@token\@onedot}
\def\@onedot{\ifx\@let@token.\else.\null\fi\xspace}

\def\etal{\emph{et al}\onedot}
\makeatother

\definecolor{blue}{RGB}{0,50,200}

\begin{document}

\title{Taking Training Seriously:\\
Human Guidance and  Management-Based\\
Regulation of Artificial Intelligence}

\author{\IEEEauthorblockN{Colton R. Crum$^{*}$}
\IEEEauthorblockA{\textit{Computer Science and Engineering} \\
\textit{University of Notre Dame}\\
Notre Dame, USA \\
ccrum@nd.edu}
\and
\IEEEauthorblockN{Cary Coglianese$^{*}$}
\IEEEauthorblockA{\textit{Professor of Law and Political Science} \\
\textit{University of Pennsylvania}\\
Philadelphia, USA \\
cary\_coglianese@law.upenn.edu}
}

\maketitle
{\def\thefootnote{}\footnotetext{$^{*}$ denotes equal contribution.}}
\begin{abstract}
Fervent calls for more robust governance of the harms associated with artificial intelligence (AI) are leading to the adoption around the world of what regulatory scholars have called a management-based approach to regulation. Recent initiatives in the United States and Europe, as well as the adoption of major self-regulatory standards by the International Organization for Standardization, share in common a core management-based paradigm. These management-based initiatives seek to motivate an increase in human oversight of how AI tools are trained and developed. Refinements and systematization of human-guided training techniques will thus be needed to fit within this emerging era of management-based regulatory paradigm. If taken seriously, human-guided training can alleviate some of the technical and ethical pressures on AI, boosting AI performance with human intuition as well as better addressing the needs for fairness and effective explainability. In this paper, we discuss the connection between the emerging management-based regulatory frameworks governing AI and the need for human oversight during training. We broadly cover some of the technical components involved in human-guided training and then argue that the kinds of high-stakes use cases for AI that appear of most concern to regulators should lean more on human-guided training than on data-only training. We hope to foster a discussion between legal scholars and computer scientists involving how to govern a domain of technology that is vast, heterogenous, and dynamic in its applications and risks.

\end{abstract}

\section{Introduction}
\label{sec:introduction}
Around the world, the rapid use and deployment of artificial intelligence (AI), especially in scenarios which could present significant adverse safety or human rights impacts, has raised fervent calls for regulation. In the face of the blistering pace of AI development, real regulatory movement is starting to take place in both the European Union (EU) and the United States, as well as at international standard-setting bodies. In March 2024, the EU adopted an AI Act that it had initially proposed in 2021—a law which now governs the use and development of AI by both public and private entities in Europe \cite{euAIact}. Within the United States, President Joseph Biden in October 2023 issued Executive Order 14,110 which outlines AI initiatives across federal agencies and provides a framework for both governing the use of AI by these agencies as well as prompting them to develop policies to guide or regulate the use of AI by private entities \cite{joseph_r_biden_jr_executive_2023, young2024memo}. And in 2023, one of the leading international standard-setting organizations—the International Organization for Standardization (ISO)—issued two major risk management standards on the use of AI: ISO 23,894 and ISO 42,001 \cite{ISO23894, ISO42001}.

Within these major regulatory developments, there appears to be consistent concern about AI’s use within systems deemed to present high risks, such as in medicine and with autonomous vehicles. Within these domains, simply training AI models without human oversight will become impermissible. The EU AI Act, for example, specifically mandates the incorporation of explicit human oversight into AI training in certain systems \cite{euAIact}. Executive Order 14,110 similarly calls for “careful oversight” of the development and application of AI, while also directing the federal government to refine standards and guidelines for “appropriate procedures and processes” that are deemed necessary “to help ensure the development of safe, secure, and trustworthy AI systems \cite{joseph_r_biden_jr_executive_2023}.'' The ISO now directs organizations that develop and use AI to establish management systems that, among other things, “shall define a process for assessing the potential consequences for individuals and groups of individuals, or both, and societies that can result from the development, provision or use of AI systems \cite{ISO42001}.''

What these approaches have in common is a “management-based” regulatory framework that contemplates human oversight of the design, deployment, and use of AI systems. Rather than commanding the use of any particular model or method, a management-based approach to AI governance instead calls upon those who develop and deploy AI systems to follow a process for identifying risks, anticipating undesirable consequences, and establish and follow procedures for AI system validation and auditing \cite{coglianese2010management, coglianese2003management}.

With the human oversight demanded of a range of emerging management-based AI standards, the techniques of human-guided training of AI models are likely to prove pivotal in the years ahead. Human-guided training may also provide the breadth necessary to handle some of the common challenges that modern AI faces, such as lack of interpretability, misalignment with human intuition, and output explainability. In this paper, we highlight some of the key advantages of human-guided training for purposes of regulatory compliance and argue for their staying power with systems that pose potentially high impacts on safety or human rights. We note how human guidance within the training of AI tools based on supervised learning can be applied at the input, architectural, and loss function level, and we broadly highlight other techniques of human guidance for reinforcement and unsupervised learning. Next, we argue that some of the common regulatory difficulties associated with AI can be alleviated with human-guided approaches to AI models dependent on any type of machine learning. Finally, we offer some strategies for both firms and regulators in finding a middle ground that will satisfy technical and legal concerns. We seek to facilitate discussion involving human oversight during model training, highlighting the importance of taking training seriously while also discussing some of the challenges it might present.
\section{Emerging Management-based AI Regulation}
For the past few years, the debate on how to regulate AI, if at all, has proven to be contentious. Calls to regulate AI have included the creation of dedicated AI regulatory bodies \cite{lieu2023} and the imposition of insurance requirements \cite{lior2022insuring}, tort liability \cite{calo2015robotics, crootof103international, crootof2015war, karnow2016application, marchant52coming}, or even outright bans \cite{yudkowsky2023}. Notwithstanding the wide array of regulatory options, the most advanced, tangible regulatory progress appears to be settling on a management-based approach to AI governance because of its flexibility and ability to apply to a wide range of AI use cases. This management-based approach is reflected in recent regulatory developments in the United States and Europe, as well as in the standards issued by major nongovernmental standards-setting bodies, such as the ISO.

\subsection{U.S. Regulatory Developments}
The U.S. Congress has adopted to date two pieces of legislation on AI: the AI in Government Act of 2020 and the National AI Initiative Act (NAIIA) of 2020. These pieces of legislation established entities in the federal government to begin to study and monitor uses of AI, but they do little themselves to regulate private or public sector use of AI. As a result, numerous additional pieces of legislation have been proposed. In September 2023, for example, Senators Wyden, Booker, and Clarke introduced the Algorithmic Accountability Act of 2023 which would require firms to assess the impacts of AI systems and products. This bill would require the Federal Trade Commission (FTC) to establish standardized impact reporting methods to be followed in overseeing the impacts of critical automated decision-making. 

Although Congress has not yet adopted any comprehensive AI legislation, federal agencies throughout the executive branch have responded in the last several years with dozens of measures that take the form of binding administrative actions, various guidance documents, and plans and reports. Specific regulatory agencies have already detailed AI-related guidance or even regulations, such as those issued by the National Highway Traffic and Safety Administration (NHTSA) with respect to the use of automated driving systems or by the Food and Drug Administration (FDA) with respect to AI-assisted medical devices. At the end of October 2023, President Biden issued Executive Order 14,110 \cite{joseph_r_biden_jr_executive_2023}, which details a broad set of principles for AI regulation and calls upon federal agencies in the executive branch to do more to develop flexible forms of rules related to public and private sector uses of AI.\footnote[1]{\url{https://www.whitehouse.gov/briefing-room/presidential-actions/2023/10/30/executive-order-on-the-safe-secure-and-trustworthy-development-and-use-of-artificial-intelligence/}} In conjunction with this executive order, the White House also issued from Office of Management and Budget (OMB) a memorandum to all federal executive agencies that sets forth standards for public sector use of AI \cite{young2024memo}. This memorandum directs federal agencies to ensure that certain types of “safety-impacting” or “rights-impacting” uses of AI are to be subjected to additional layers of auditing, testing, and monitoring.\footnote[2]{\url{https://www.whitehouse.gov/wp-content/uploads/2024/03/M-24-10-Advancing-Governance-Innovation-and-Risk-Management-for-Agency-Use-of-Artificial-Intelligence.pdf}}

\subsection{EU AI Act}
In April 2021, the EU Commission published a proposal to regulate AI in the EU \cite{euAIact}. This proposed AI Act resulted in a provisional agreement on EU legislation in December 2023, and finally passing in March of 2024.\footnote[3]{\url{https://www.europarl.europa.eu/doceo/document/TA-9-2024-0138_EN.html}} 
The EU AI Act aims towards regulating all aspects of AI in an effort to promote the safety and fairness of AI-based products sold in the European market as well as to ensure that public and private sector uses of AI in Europe are respectful of fundamental human rights. More specifically, the Act applies additional regulatory scrutiny for AI uses that are deemed to pose high risks or that involve high-impact general-purpose AI that could contribute to larger societal risks. In instances where the risk of using AI exceeds what the law deems a tolerable threshold, AI use might well warrant a full ban, such as uses related to cognitive behavior manipulation, social scoring, and predictive inference on sensitive information like sexual orientation or religion. Other uses deemed to be high-risk—defined as uses posing a significant, foreseeable, or severe risk of harm to health, safety, or fundamental rights—a management-based regimen will be required. AI models for these uses will be subject to pre-market testing, reporting measures, and model auditing. A newly established AI Office within the Commission will oversee the most advanced AI models, set risk-specific regulatory standards, and enforce penalties and bans.

\subsection{ISO Standards}
The ISO is one of the foremost international standard-setting bodies, having for decades established a wide range of product- and process-related technical standards. Although hundreds of other private standards on AI have emerged in recent years \cite{marchant2022soft}, the ISO is one of the most prominent standard-setting bodies to have weighed in on AI governance. In February 2023, ISO adopted standard 23,894 on the risk management of AI systems which offers a framework for organizations that develop and use AI tools. It calls on these organizations to “implement a risk-based approach to identifying, assessing, and understanding the AI risks in which they are exposed and take appropriate treatment measures according to the level of risk \cite{ISO23894}.'' Later, in December 2023, ISO adopted standard 42,001 on the establishment of organizational management systems to address the risks from the use of AI. That standard, and the accompanying guidance on implementing AI controls, outlines a systematic framework for identifying and assessing AI risks and then monitoring, measuring, and evaluating organizations’ application of that framework. It calls for documentation and management review, indicating that an organization that designs or deploys AI “shall continually improve the suitability, adequacy, and effectiveness of [its] AI management system \cite{ISO42001}.”

\subsection{Management-Based Oversight of AI}
Management-based regulation is a type of regulation that requires regulated entities to engage in internal managerial steps to identify risks, establish measures to reduce or control them, and then audit to ensure both that the selected measures are followed as well as that risks are kept under control \cite{coglianese2010management, coglianese2003management}. It is used widely in a variety of contexts, from aviation safety, chemical accident avoidance, and prevention of foodborne illnesses. As has been noted in a U.S. National Academies of Sciences study from 2018, management-based regulation can be an appropriate strategy in high-hazard contexts where the source of the underlying risks are varied and methods for defining, monitoring, or enforcing performance are difficult or deficient \cite{national2018designing}.

As is apparent from the emerging U.S., EU, and ISO standards, management-based governance now appears to be the approach that will apply to AI. This approach is understandable given the heterogeneous nature of AI as a suite of algorithmic tools, the varied uses to which it is put, and the wide range of possible problems that it could create (or ameliorate) through its application \cite{coglianese2023regulating}. The general approach reflected in the U.S., EU, and ISO standards consists of two core structures relevant to the data scientists and firms that develop AI models: (i) a threshold for heightened scrutiny, and (ii) a set of standards for the systematic human-driven oversight of those models that surpass the relevant threshold. The first of these two structures—the threshold for scrutiny—provides a basis for distinguishing applications of AI along a spectrum from those relatively benign uses that do not merit much, if any, human oversight to those that, if not needing to be banned altogether, necessitate heightened, rigorous oversight. We focus here on uses in the latter end of this spectrum, where the emerging regulatory regime will dictate the application of management-based interventions as a form of human oversight to those systems.

Also evident from these varied sources of emerging standards is a common treatment of AI deemed to pose heightened safety or rights-related risks to consumers or to the public. These uses will be expected to be developed in accordance with a structured management system at all stages of the AI pipeline. Within the U.S., for example, the executive order and the OMB memorandum outline a series of management-based steps that will be required—and these steps aim to ensure that humans will serve as a “checks and balances” to AI systems, especially post-deployment. The OMB memorandum, for example, states that “[a]gencies must ensure there is sufficient training, assessment, and oversight for operators of the AI to interpret and act on the AI’s output, combat any human-machine teaming issues (such as automation bias), and ensure the human-based components of the system effectively manage risks from the use of AI.” 

The need for human oversight after launching an AI-based tool is palpable and certainly reflective in the emerging management-based standards’ calls for evaluation and continuous monitoring of AI systems, especially for applications in high-risk domains such as the automotive and aviation industry. Even without AI, technologies in these domains are already subject to regulatory controls that, among other things, aim to ensure that they are carefully monitored in use to identify acute problems, reduce the risks of catastrophic failure, and make necessary technical changes for public safety. With AI, there will also exist a need for such testing and in-use monitoring. But what is too little appreciated, and still perhaps most important, is the need for careful and direct human involvement into the creation and training of AI.
\section{Management-based Regulation and Human-Guided Training}

The EU AI Act speaks most directly to the need for careful human management of the creation of high-risk AI systems. In Section 2 Article 14, the Act states that “[h]uman oversight shall be ensured through either one or all of the following measures: (a) identified and built, when technically feasible, into the high-risk AI system by the provider before it is placed on the market or put into service \cite{euAIact}.” This provision invites human oversight not only at the evaluation or deployment stage, but directly “built” into the model. This is important because the role for humans in the AI creation process is markedly different from their role in more traditional, even if still complex, industrial or product manufacturing processes. 

In the past, techniques to build intelligence into computers sought hard coded, rule-based reasoning through a series of conditional statements. In early digital systems, features were hand-crafted directly into a computer model, a laborious and exhausting task that limited its use to the defined task. Slight variations in the input data could render models obsolete. Over time, developments in machine learning architectures and training techniques have allowed models to learn on their own, without the tedious human guidance that was once required. The “self-learning” component allows models to build their own solutions and representations to solve requested tasks. This technical shift has allowed AI to scale to larger and more complex problems, spanning across a broad range of tasks, including those involved in processing audio, vision, and language inputs. 
AI’s ability to solve complex tasks from the observation of training data has spawned both admiration and concern. Although machine learning can exceed human performance on certain tasks, its self-learning properties can also prevent humans from seeing exactly how certain patterns or representations were learned or forecasts were generated, which makes for a qualitatively different type of regulatory challenge. Outcomes are not intuitively explainable and the likelihood of unintended consequences seems ever-present. The ability of AI tools to be developed with minimal explicit human oversight, then, is arguably both the “feature” as well as the “bug” of machine learning.
But this property of machine learning means that model training and assembly is one of the most pivotal points of risk management. This stage—when the model is being formed and when it is learning—is arguably the only time humans can potentially have their hand in overseeing the algorithm’s self-learning properties. An important goal of management-based regulation should be to ensure that the algorithm itself is well-managed, which means that greater regulatory emphasis and scrutiny is likely to encourage AI developers not merely to develop after-the-fact testing, evaluation, and disclosure but to approach the process of training their AI models with care.
For high-risk uses of AI, even just the disclosure of basic training details will likely not suffice if the training was largely data-driven. For example, with AI that aids physicians in classifying cancer—or with any other system that aids humans in making potentially life-consequential decisions—responsible management of AI will not allow the solution space to be dictated merely by a process through which the algorithm learn patterns within the training data all on its own. Rather than just taking a machine-learning model and applying it to a large dataset and hoping for the best, the better practice will be to have humans closely oversee and interrogate the training itself. 

The EU AI Act has already noted that human oversight must be “built, when technically feasible, into the high-risk AI system by the provider.” For these high-risk systems, the regulatory direction will be for humans to have some hand in guiding the self-learning components of AI. Despite the existence of some current industry practices that simply rely on mammoth amounts of training data and the brute force of machine-learning computing, humans should go beyond merely testing and monitoring after the AI model has been trained for uses where the potential consequences are significant. It is important to take training seriously and consider pursuing research and data analytic techniques 
aimed at combining the strengths of machine learning and humans together through models that are human-guided.

\section{Techniques of Human-Guided Training}
\begin{figure*}[t]
  \centering
  \includegraphics[width=\linewidth]{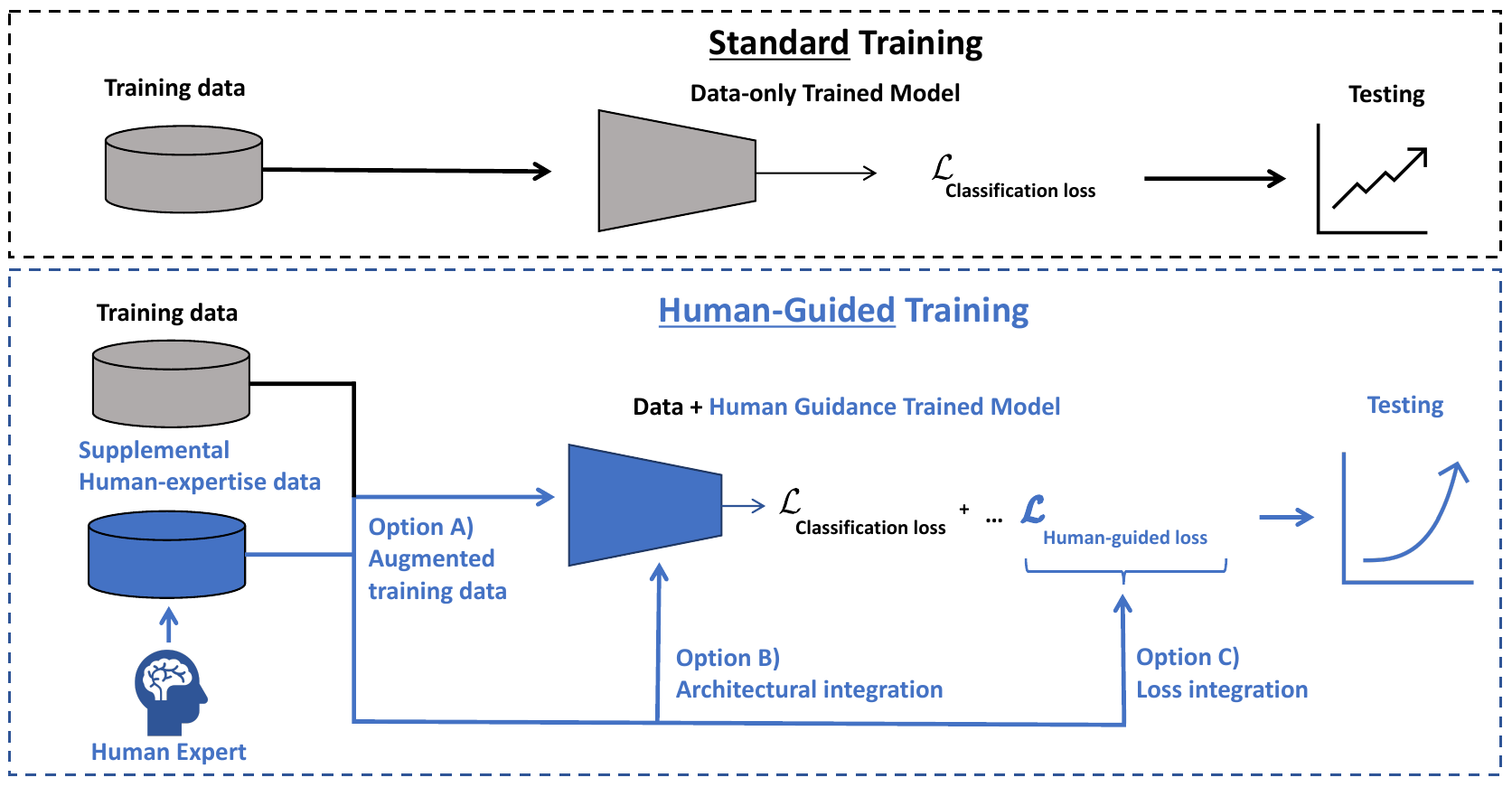}
  \caption{Examples of standard (top) and \textcolor{blue}{human-guided training configurations (bottom)} for \text{supervised learning}. Standard training involves learning only from the training data \emph{without} any human oversight. \textcolor{blue}{Human-guided training involves collecting supplemental human-expertise data and incorporating this informational oversight into model training through three options: (a) augmented training data, (b) architectural integration, or (c) loss integration.}}
  \label{fig:human-training}
\end{figure*}
What might these techniques of human-guided training entail? Here we illustrate research efforts into human-guided training aimed at improving the explainability and interpretability of AI. We focus, for sake of presentation, on the supervised training of AI tools to perform vision-related tasks and the challenge of explainability that they pose. Although we illustrate practices of human-guided training by reference on vision-related tasks, we do not mean to imply that this is the only domain in which human guidance will be valuable. Quite the contrary, other forms of AI will almost surely benefit as well, including those related to natural language processing (NLP), recommendation systems, audio-to-text speech, games, and generative tooling (audio, visual, or textual). To show how the notion of human guidance applies more broadly, we briefly describe several other human-guided training techniques that extend to forms of AI other than supervised learning, such as in reinforcement and unsupervised learning. Further research on best practices for training AI systems across a wide range of tasks and with respect to a range of other risks and concerns will be needed.

\subsection{The Explainability Challenge}
Since the invention of AI, humans have sought understanding and interpretation of its outputs. For rudimentary statistical techniques, such as decision trees, linear regression, and rule sets, the process of interpretation can be fairly straightforward and intuitive. With the advent of more technically sophisticated machine learning, however, it has become more challenging to explain model outputs. Support vector machines (SVM), multi-layer perceptrons, and neural networks introduce many learning non-linearities that increase performance capabilities but limit conventional model interpretation methods. As a result, research in explainable AI (xAI) and model “interpretability” has sought techniques to inform humans of model outputs and understanding of features text, images, signal, or tabular data
\cite{tjoa2020survey, linardatos2020explainable}. Model explainability has been approximated through techniques involving clustering \cite{nguyen2016multifaceted}, neuron activations \cite{zhou2016learning}, principle component analysis \cite{olah2017feature}, and gradients \cite{gradcam}. Auditing models for larger concepts like “trustworthiness” has been developed through trust scores \cite{jiang2018trust, wong2020much}, Trust Matrices and Densities \cite{hryniowski2020does}, and Chain of Trust \cite{toreini2020relationship}.

Vision-related tasks are sensibly dominated by visualizations. By far the most common visualization for neural networks are through saliency maps, which are methods that aim to assess “where” in the input image the model was “looking” when it made its decision (classification). Saliency maps can use internal features directly via neuron activations \cite{zhou2016learning}, gradients \cite{gradcam}, or entirely black-box \cite{petsiuk2018rise, LIME, SHAP}. These methods offer a quick “sanity check” compared to human intuition, and they are used qualitatively to verify the model’s understanding of the task. Many saliency map methods are used in human-guided training techniques.

The steady increase in AI’s performance has led to its deployment on human-level tasks, whether working alongside humans or even substituting for them. In situations of “human-AI teaming,” where the AI is operating within a human environment, effective communication between both the humans and the AI system is necessary to operate successfully. Successful human-AI teaming typically involves supplemental model information (i.e., saliency maps, explainability measures, added captions) besides traditional model outputs as a means of more comprehensively conveying model information to the human. However, achieving successful complementary performance—that is, having the AI and humans work better together than each can separately—has proven to be difficult in practice. Decreased overall performance has been observed within clinical treatments \cite{jacobs2021machine}, and even lackluster performance within human understanding of AI-based explanations \cite{bansal2021does}. Physiological factors such as perceived confidence within the AI’s ability can contribute overreliance, where the human overlies on the AI’s output.
Moreover, although AI has demonstrated impressive performance across many challenging tasks, AI systems do not always follow human intuition. Vision models can output to humans in classification tasks but latch onto extraneous features located in the inputs that are correlated with class labels. During standard or conventional training which only focuses on model accuracy, the AI is simply instructed to minimize the training error and receives no instruction on “where to look” within a given image. This has led to misalignments between the model and the human visual salience, which inevitably can erode model trust and weaken the complementary AI-human team’s performance \cite{jacobs2021machine, bansal2021does,linsley2017visual}.

\subsection{Human-Guided Training within Supervised Learning}
Human-guided training seeks to combine explainability with human oversight, attempting to incorporate this information during training. Rather than simply setting up a machine-learning algorithm and telling it to process a large volume of data, human-guided training uses expertise or domain-specific knowledge as supplemental information to guide the model during training. Models developed with human-guided training have been shown to improve explainability and interpretability \cite{crum2023explain, linsley2018learning, fel2022harmonizing}, converging faster to a solution space and improving generalization and overall model performance \cite{boyd2021cyborg, boyd2023patch, linsley2018learning}.

Information from human experts can be collected through reaction times \cite{huang2023measuring, fel2022harmonizing}, eye-tracking \cite{boyd2023patch, czajka2019domain}, written annotations \cite{boyd2021cyborg,boyd2022human}, and games \cite{linsley2018learning}.

After collection, there are three broad ways for introducing human guidance into the training process for supervised learning. These include incorporating explicit human involvement by: (a) augmenting the training data, (b) the alignment of neurons or other internal model architectures, and (c) integration into the loss function (see Fig. \ref{fig:human-training} for an example of human-guided training for supervised learning).

\subsubsection{Augmented training data} The most straightforward way of incorporating human intelligence into the model is through its training data (Option A in Fig. \ref{fig:human-training}). Models are only capable of extracting information from which it's given. Therefore, providing high-quality, biased-free training samples is extremely important to a model’s performance and capabilities. However, state-of-the-art models are demonstratively slow learners, and requiring several magnitudes more data to learn simple tasks compared to humans. In some instances, the amount of training data required for a model to effectively learn a task means engineers must make a trade off between data quantity and quality. This makes training and collecting on purely domain-expert data sometimes improbable depending on the task. However, there are some techniques to augment training data and embed human expertise into the training data to encourage human-aided learning. Boyd \etal utilized blurring to discourage the model from non-salient regions of the input image provided by human annotators \cite{boyd2022human}.

\subsubsection{Architectural integration} Research has sought to understand human attention and artificially replicate these processes (Option B in Fig. \ref{fig:human-training}). Inspired by how humans use global and local features to “encode” visual stimuli, Linsley \etal proposed a global-and-local attention (GALA) block that mimics parallel attention pathways from human-supplied saliency maps \cite{linsley2018learning}. Several other bodies of work aim to appropriately capture model “attention” and align internal mechanisms with these processes \cite{fel2022harmonizing}. However, these internal attention mechanisms usually require invasive, often architectural-specific changes to the model.

\subsubsection{Loss integration} A more classical and less invasive way of incorporating human-salient information into model training is through the loss function (Option C in Fig. \ref{fig:human-training}) \cite{huang2023measuring}, \cite{boyd2021cyborg}, \cite{piland2023model}, \cite{ismail2021improving}. Rather, it can be applied to a wide variety of models with little computational overhead. The loss function is the overall training mechanism used to instruct model, comprising of an error to which the model is attempting to reduce, or minimize during training. Largely, this is based off of performance or accuracy ($\mathcal{L}_\text{classification}$ loss in Fig. \ref{fig:human-training}). Human-guided models can be integrated into the loss by adding a second penalty to which the model is attempting to satisfy ($\mathcal{L}_\text{Human-guided}$ loss in Fig. \ref{fig:human-training}). In many of these cases, the model is trying to perform accurately while simultaneously, for example, aligning itself with human intuition or human saliency maps \cite{boyd2021cyborg}, \cite{ismail2021improving}. Boyd \etal introduced ConveY Brain Oversight to Raise Generalization (CYBORG), a method that directly compares the models Class Activation Mapping (CAM) with the human-salient annotation \cite{boyd2021cyborg}. This allows the model to not only learn its own representations of the training data, but also consider information that humans used as well (saliency maps).

\subsection{Other Types of Human-Guided Training}
Our analysis of human-guided training thus far has focused largely on supervised learning contexts, where humans provide the ground truth labels for which the AI learns the desired task. However, human-guided training can be incorporated into other conventional training paradigms without ground truth labels, such as reinforcement or unsupervised learning. Within reinforcement learning, human feedback has been used to modify an AI agent's reward \cite{pilarski2011online, knox2012reinforcement}, or even to inform the policy directly \cite{griffith2013policy, knox2012reinforcement}. For unsupervised learning, human-guidance involves, at a minimum, taking the training data more seriously. Any type of generative or foundational AI requires the careful assembly of training samples, which often require widespread crowdsourcing platforms to clean and assemble. Unsupervised models can be guided towards a human preferred solution space using domain-specific, expertise curated samples \cite{hendrycks2021cuad}, or even direct processing techniques such as task-specific augmentations can likely be applied \cite{boyd2022human} (Option C in Fig. \ref{fig:human-training}).

\section{Advantages of Human-Guided Training}
Which of these three ways of incorporating human guidance into model training will be most effective is likely to vary depending on the use case and its larger context. Factors that will likely matter for the success of human guidance include the availability of sufficient training data, the feasibility of providing human-annotations, and the existence of sufficient human expertise. Still, the overarching point is that human oversight within the training processes can alleviate some of the downsides associated with the self-learning properties of AI and can more effectively guide their application in high-hazard contexts. Human-guided training can help, for example, in (a) aligning models with human-salient information, (b) creating more human-explainable or interpretable outputs, or (c) encouraging a shared-nature between human and machine intelligence. We can summarize several key advantages of human-guided training, especially in an era of management-based AI regulation. 

\subsection{AI-Autonomous Decisions}
Just as management-based regulation is being adopted to cover a wide span of AI use cases, human-guided training techniques can be applied in a range of scenarios. The scenarios that appear to raise the greatest public concern, ceteris paribus, are those where AI systems replace humans entirely—or so-called human-out-of-the-loop uses. When AI systems are used to replace human decision-making—say, by acting autonomously on behalf of an employer—management-based regulation will still call for attentiveness to risks and the use of risk-mitigation procedures and techniques. It is just that, in these circumstances, the key venue for human attentiveness must be, by necessity, directed at the model training. Yet even in instances where AI merely informs human decisions rather than substitutes for them, human-guided training can lead to systems that will better communicate or ``explain'' outputs with the human decision maker \cite{crum2023explain, linsley2018learning}.

\subsection{Regulatory Approvals}
Another major advantage of human-guided training is its potential for facilitating regulatory approvals. The EU AI Act suggests that regulators may insist on pre-market evaluation measures for certain kinds of AI systems. Although the precise requirements remain nascent, human-guided training is likely to figure into what regulators will come to expect. At a minimum, well-documented evidence of such guidance during training seems likely to offer the prospect for expediting regulatory approvals. Accelerated regulatory approval is common in the pharmaceutical industry, where the FDA offers a fast track review of drugs intended to treat urgent and serious medical conditions. Fast track drugs display early advantages over current and available treatments, and they can bypass several otherwise regulatory requirements. AI displaying “early” advantages over state-of-the-art (SOTA) performance may not directly apply, but the use of human oversight during training could allow for waivers, model-specific exemptions, or other regulatory exceptions \cite{coglianese2021unrules}. Top-down model scrutiny might be relaxed in some circumstances in the face of robust documentation of human-led oversight during training of model-specific architectures, domain-specific challenges, or environmental interventions.

\subsection{Improved Outcomes}
Models based on human-guided training may potentially allow AI deployment when its would otherwise be deemed unacceptable. Various legislative proposals, including the EU AI Act, take the position that for certain AI uses, machine-based failures cannot be tolerated and thus the use of AI should be outright banned. Yet if the risks associated with the use of AI in some contexts or for some uses may be too great, then perhaps the risks associated with continued reliance solely on human decision-making ought to be considered suspect as well. If it is possible to use AI to improve on the status quo—through models based on human-guided training—then blunt rules that foreclose such innovation can leave society worse off. By introducing humans into the training process, some AI-specific risks can be reduced, which may allow an acceptable threshold of safety to be achieved that would otherwise remain unmet.

\subsection{Compatibility with Management-based Regulation}
Finally, to highlight a major theme of this paper, when the major forms of AI regulation rely upon a management-based approach, the need for human guidance during training will become increasingly expected. Management-based approaches necessarily call for greater reliance on human processes, internal documentation and reporting, and testing protocols to govern \cite{coglianese2023regulating}. Because the design of the model itself will play a pivotal role in the risks it poses to consumers and the public, any entity that takes seriously its obligations under management-based regulation must also take seriously the need for an iterative and multifaceted training process.
\section{Limitations}
Despite the benefits we have highlighted of human-guided training, we should not be blind to their potential limitations. Models based on human-guided training will, by necessity, be potentially susceptible to some of the limitations of human oversight. This will include, in the first instance, the direct costs of acquiring human-salient information for each training sample. When the additional costs in time and money are substantial, developers may cut corners and the full potential of human guidance to training may never be realized. If costs limit the quantity of available human-salient training data, for example, this will likely limit the potential for human guidance to offset the dangers of relying solely on data-based training processes. Although ongoing research seeks to identify methods for exploiting the capacity of existing human-annotations \cite{crum2023teaching}, acquiring sufficient quantities of training samples can prove to be challenging. For some tasks and domains, a large reduction in training data could cripple an AI system’s performance and reducing its usefulness.
Other limitations stemming from human cognition or motivation can impede the viability of human-guided training to overcome all problems with AI systems \cite{coglianese2021algorithm}. Since human-guided models reduce the self-learning component of AI, the burden of error is distributed between the human and the AI. For example, errors committed by a board-certified radiologist will propagate into the model, encouraging it to make similar errors on similar samples. Consequently, determining whether an AI error originated from the human-supplied data (radiologist) or the AI itself could prove to be challenging. However, some human-guided models have shown improved interpretability, which may reduce these instances of overall error.

In some sense, the human-expert’s cognitive ability may become a ceiling for human-guided AI. Once AI’s performance exceeds the human, it may be problematic for the human to inform the AI, and this could motivate the reverse scenario: the AI informs the human. In these contexts, the AI may not benefit any more from human-salient information (such as where the radiologist was looking on each scan), but from physiological information. Human physiological-information such as reaction times have proven to regularize model output \cite{huang2023measuring}. Although currently these scenarios seldom exist, finding the intersection of useful knowledge and its modality (i.e., visual, psychological, physiological) between AI and human teams will be a challenge in the near future.
Finally, although explainability methods are widely used in the community, there remains debate over the faithfulness of these explainability measures. Kindermans and Hooker \etal showed that saliency methods can be maliciously distorted by simple data preprocessing steps \cite{kindermans2019reliability}. Other researchers have found similar findings, including architectural nuances and the affect of layer-wise randomizations \cite{adebayo2018sanity, binder2023shortcomings}. Reducing the efficacy of AI explainability can reduce the fruitfulness of human-guided models as many strategies are built upon explainability measures. Regardless of shortcomings of specific explainability methods, understanding model outputs and conveying their meaning to human operators remains an important and vital research direction for AI \cite{lipton2018mythos}. Like all areas of AI, more research will be needed to expand these explainability tooling and to keep up pace with architecture advances and new training methods.

\section{Conclusion}
In this paper, we have discussed regulatory approaches emerging from U.S. and EU authorities as well as international standard-setting organizations. These approaches seem to be converging on a structure that imposes a threshold for heightened regulatory scrutiny. Whether that threshold is described in terms of “high-risk” or “safety- or rights-impacting” consequences, once it is exceeded the kind of regulatory scrutiny will largely take the form of demonstrated management-based interventions, such as impact assessments and auditing. Complying with these management-based requirements will necessarily involve human oversight or guidance, especially within the training process. We have argued that the training of a machine-learning algorithm is a crucial step in the risk management of an AI system, and we have suggested three main avenues for responsible human intervention into the training process. When human experts can meaningfully guide the training of AI through the careful assembly of training data, integration directly into the model architecture, or incorporation into the loss function, they may have a better potential to address the concerns that are animating the development of AI regulation. Especially in high-hazard domains or for high-risk uses, neither system developers nor regulators should be satisfied with merely letting the algorithm train itself in an uninterrogated manner. The responsible design of AI models in such circumstances necessitates taking training seriously.  

\small
\bibliographystyle{IEEEtran}
\bibliography{egbib}

\end{document}